\documentclass[3p, times,merge]{elsarticle}
\journal{Computer Physics Communications}
\usepackage[final,notref,notcite]{showkeys}
\usepackage{amsmath}
\usepackage{amssymb}
\usepackage{bm}
\usepackage[ISO=false]{diffcoeff}[=v4]
\usepackage{listings}
\usepackage{graphicx}
\usepackage[utf8]{inputenc}
\usepackage{xcolor}
\usepackage{calc}
\usepackage{array}

\definecolor{mydarkgray}{gray}{0.2}

\usepackage{algorithmicx}
\algrenewcommand{\algorithmiccomment}[1]{{\color{darkgray}\small\# \textit{#1}}}
\usepackage[plain]{algorithm}

\usepackage{tikz}
\usetikzlibrary{calc, arrows.meta}

\newcommand{\ess}{\operatorname{ESS}}
\newcommand{\re}{\operatorname{Re}}

\newcommand{\tr}{\operatorname{Tr}}
\newcommand{\sign}{\operatorname{sign}}
\newcommand{\bz}{\bm{z}}
\newcommand{\bphi}{\bm{\phi}}
\newcommand{\btheta}{\bm{\theta}}
\newcommand{\vphi}{\varphi}
\newcommand{\bvphi}{\bm{\vphi}}
\newcommand{\avg}[1]{\left\langle #1 \right\rangle}

\newcommand{\pytorch}{\texttt{PyTorch} }
\newcommand{\reinf}{RE}
\newcommand{\rt}{r.t.}

\newcommand{\pacpt}{p_{a}}

\newcommand{\var}[1]{\operatorname{var}\left[#1\right]}
\newcommand{\E}[1]{E\left[#1\right]}
\newcommand{\g}{\mathbf{g}}
\newcommand{\grt}{\g_{rt}}
\newcommand{\gri}{\g_{RE}}
\newcommand{\qpt}{q(\bphi|\btheta)}
\diffdef{}
{
    op-symbol = \mathrm{d},
    *-op-left = true,
    op-order-sep = 0 mu,
}

\diffdef{p}
{
    left-delim = \left .,
    right-delim = \right|,
    subscr-nudge = 1 mu
}
\begin{document}

\begin{frontmatter}

\title{Training normalizing flows with computationally  intensive target probability distributions}
\author[iis]{Piotr Białas}
\author[ift]{Piotr Korcyl}
\author[ift]{Tomasz Stebel}

\address[iis]{Institute of Applied Computer Science, Jagiellonian University, ul.~Łojasiewicza 11, 30-348 Kraków Poland}
\address[ift]{Institute of Theoretical Physics, Jagiellonian University, ul.~Łojasiewicza 11, 30-348 Kraków Poland}

\begin{abstract}
 
Machine learning techniques, in particular the so-called normalizing flows,  are becoming increasingly popular in the context of Monte Carlo simulations as they can effectively approximate target probability distributions. In the case of lattice field theories (LFT) the target distribution is given by the exponential of the action.
The common loss function's gradient estimator based on the "reparametrization trick" requires the calculation of the derivative of the action with respect to the fields.  This can present a significant computational cost for complicated, non-local actions like {\em e.g.} fermionic action in QCD. In this contribution, we propose an estimator for  normalizing flows based on the REINFORCE algorithm that avoids this issue. We apply it to two dimensional Schwinger model with Wilson fermions at criticality and show that it is up to ten times faster in terms of the wall-clock time as well as requiring up to $30\%$ less memory than the reparameterization trick estimator.    It is also more numerically stable allowing for single precision calculations and the use of half-float tensor cores. We present an in-depth analysis of the origins of those improvements. We believe that these benefits will appear also outside the realm of the LFT, in each case where the target probability distribution is computationally intensive. 
\end{abstract}

\end{frontmatter}

\section{Introduction}

Monte Carlo simulations remain a very important computational tool in many areas ranging from social sciences, Bayesian data analysis, and inference to physics. In many cases to generate samples from a given target distribution one resorts to the construction of an associated Markov chain of consecutive proposals \cite{metropolis, *Hastings}. The only limiting factor of the approach is the statistical uncertainty which directly depends on the number of statistically independent configurations. Hence, the effectiveness of any such simulation algorithm can be linked to its autocorrelation time which quantifies how many configurations are produced before a new, statistically independent configuration appears. For systems close to phase transitions the increasing autocorrelation time, a phenomenon called critical slowing down, is usually the main factor which limits the statistical precision of outputs. 

The recent interest in machine learning techniques has offered possible ways of dealing with this problem. 
Ref.~\cite{PhysRevD.100.034515} proposed normalizing flows based on neural networks as a mechanism for generating independent configurations in lattice field theories (LFT) which can be used as proposals in the construction of the Markov chain. The new algorithm was hence called Neural Markov Chain Monte Carlo (NMCMC).  For discrete statistical systems like e.g. the Ising model, autoregressive neural networks were used in the NMCMC sampling algorithm~\cite{VANPRL,PhysRevE.101.023304,Bialas:2021bei,Bialas:2022qbs,Bialas:2022bdl,Bialas:2023fjz}. Once the neural network is sufficiently well trained, one indeed finds that autocorrelation times are significantly reduced as was demonstrated in the context of the two-dimensional Ising model in Ref.~\cite{Bialas:2021bei}.

Neural networks that build up the normalizing flows have to be trained, {\em i.e.} their weights should be tuned so that the model can approximate the desired probability distribution. The standard approach for achieving this is using the stochastic gradient descent (SGD) algorithm which requires the estimation of gradients of the loss function with respect to the neural network weights. The most commonly used estimator of the gradient is based on the so-called "reparametrization trick" (\rt{}) \cite{reparameterization}. It is straightforward to implement but requires the calculation of gradients of the target probability. If this probability is given by a complex formula this may lead to severe degradation of performance.  
 
In Ref.~\cite{bialas2022gradient} we have proposed to use REINFORCE (\reinf{})  algorithm for the gradient estimator. We have shown how it can be implemented in case of reversible\footnote{All flows are reversible, what we mean here is that reverse transformation can be efficiently implemented.} normalizing flows and that it avoids calculating the derivative of the action. In there, we have applied this estimator to $\phi^4$ LFT and while it had better convergence properties, the $\phi^4$ action is very simple and did not bring out the full capabilities of this approach. The same implementation for RE as in  Ref.~\cite{bialas2022gradient} was later proposed also in Ref.~\cite{pathGradient}.  In this contribution, we apply this estimator to the case of the 2D lattice Schwinger model with Wilson fermions. The fermionic action requires the calculation of the determinant of the Dirac operator, which is represented by a large ($2 L^2\times 2L^2$  for $L\times L$ lattice) matrix, so avoiding propagating gradients through those calculations may prove beneficial.  This is also probably the simplest model with dynamical fermions so it is often used as a testing ground for algorithms that eventually can be used for lattice QCD \cite{PhialaSchwinger} making it an interesting model to study. We demonstrate that the \reinf{} estimator is significantly faster than the \rt{}, which is currently the most commonly used gradient estimator. Already at $L=12$ \reinf{} outperforms the r.t. estimator and the difference grows quickly with $L$, reaching a factor of 10 for $L=24$. In addition, we show that the \reinf{}  requires much less memory which plays a role for larger systems sizes. The code used in this paper is available at \cite{code}.

This paper is organized as follows: in section~\ref{sec:nmcmc} we present the Neural Markov Chain Monte Carlo algorithm and explain how it can be implemented in terms of normalizing flows. In section~\ref{sec:grad-estimators} we present \reinf{} and \rt{} gradient estimators that can be used to approximate the gradient of the loss function with respect to the model parameters. We show how the RE can be implemented in practice. In section~\ref{sec:schwinger} we introduce the 2D lattice Schwinger model and in section~\ref{sec:results} we present a detailed comparison of both estimators for this model. \ref{app:implementation} gives the details of the implementation. 

\section{Neural Markov Chain Monte Carlo}
\label{sec:nmcmc}

Monte Carlo methods rely on random samples generated from some {\em target} distribution $p(\bphi)$. Often, {\em e.g.} in lattice field theories, that distribution is complicated and depends on all degrees of freedom of the system, hence there are no methods for sampling this distribution directly and independently. Instead, the approach of choice is to construct the associated Markov chain of samples, giving rise to the so-called Markov Chain Monte Carlo approach. To be more precise, each step of the algorithm has two stages: in the first stage for the given configuration $\bphi_i$, a new trial configuration $\bphi_{trial}$ is proposed from the distribution $q(\bphi_{trial}|\bphi_i)$. In the second stage, the trial configuration is accepted with probability $\pacpt(\bphi_{trial}|\bphi_i)$ usually given by the Metropolis-Hastings acceptance probability \cite{metropolis, *Hastings}
\begin{equation}
    \pacpt(\bphi_{trial}|\bphi_i)=\min\left\{1,\frac{p(\bphi_{trial})}{q(\bphi_{trial}|\bphi_i)}\frac{q(\bphi_{i}|\bphi_{trial})}{p(\bphi_{i})}\right\}
\end{equation}
In order to keep the acceptance rate high, typically the configuration $\bphi_{trial}$ differs from $\bphi_i$ only on a small subset of degrees of freedom, {\em e.g.} single lattice site. 
By construction, the consecutive samples generated by the MCMC algorithm are highly correlated due to small incremental changes needed at each step. On the contrary, in the Metropolized Independent Sampling (MIS) algorithm discussed in \cite{Liu} one attempts to generate {\em independent} samples from some auxiliary distribution $q(\bphi)$ {\em i.e.}
\begin{equation}
q(\bphi_{trial}|\bphi_i) = q(\bphi_{trial})
\end{equation}
and then accept or reject it with the Metropolis-Hastings step,
\begin{equation}\label{eq:accept-reject}
    \pacpt(\bphi_{trial}|\bphi_i)=\min\left\{1,
    \frac{p(\bphi_{trial})}{q(\bphi_{trial})}
    \frac{q(\bphi_i)}{p(\bphi_i)}\right\}.
\end{equation}
The MIS algorithm also introduces autocorrelations because of non-zero rejection probability, however, they can be controlled by the similarity of the distributions $q(\bphi)$ and $p(\bphi)$, i.e. if $q(\bphi)$ is close enough to $p(\bphi)$ then the acceptance rate is close to one, and subsequently the autocorrelations can be substantially smaller then in the case of MCMC (see Ref.~\cite{Bialas:2021bei} for discussion). 

The difficulty of the MIS approach lies in the construction of the distribution $q(\bphi)$ which has to, at the same time, be as close to the target distribution $p(\bphi)$ as possible and allow for practical generation of configurations. Neural Markov Chain Monte Carlo proposes to employ machine learning techniques, notably neural networks, to {\em learn} the distribution $q(\bphi)$ \cite{PhysRevD.100.034515,VANPRL,PhysRevE.101.023304}. Hence, one assumes that $q(\bphi)$ can be represented by some appropriate model parametrized by some (very large) set of parameters $\btheta$ 
\begin{equation*}
    q(\bphi)=q(\bphi|\btheta).
\end{equation*}
The parameters $\btheta$ are tuned by minimizing a loss function that measures the difference between  $q(\bphi|\btheta)$ and target distribution $p(\bphi)$ . A natural choice for such loss function is the Kullback–Leibler divergence \cite{KL}
\begin{equation}\label{eq-KL}
    D_{KL}(q|p) = \int\dl\bphi \, q(\bphi|\btheta) \left(\log q(\bphi|\btheta)-\log p(\bphi)\right)= E[\log q(\bphi|\btheta)-\log p(\bphi)]_{q(\bphi|\btheta)},
\end{equation}
sometimes called {\em reversed} K.-L. divergence in this context because the target probability is given as the second argument. 

It often happens that the target distribution $p(\bphi)$ is only known up to a normalizing constant, i.e. we only have access to $P(\bphi)$,
\begin{equation}
    P(\bphi)= Z \cdot p(\bphi),\qquad Z= \int\dl\bphi P(\bphi).
\end{equation}
The constant $Z$ is typically called the {\em partition function}. Inserting $P$ instead of $p$ into the Kullback-Leibler divergence definition we obtain the {\em variational free energy}
\begin{equation}\label{eq:shifted-DKL}
\begin{split}
   F_{q} = \int\dl\bphi\, \qpt
   \left(\log q(\bphi|\btheta)-\log p(\bphi) -\log Z\right)
   = F + D_{KL}(q|p),
\end{split}
\end{equation}
where $F=-\log Z$ is the {\em free energy}. Since $F$ is independent of $\btheta$, minimizing $F_q$ is equivalent to minimizing the full loss function $D_{KL}$. We will use $P$ and $F_q$ instead of $p$ and $D_{KL}$ in what follows. Note that the possibility of directly estimating the free energy $F$ is one of the additional strengths of this approach since $F$ is very hard to access in the classical Monte Carlo simulation \cite{PhysRevLett.126.032001}.

Normalizing flows are a particular model that allow parametrizing trial probability distributions $q(\bphi)$ over a space of configuration with continuous degrees of freedom. It may be defined as the tuple of functions \cite{PhysRevD.100.034515,dinh2017density,9089305}
\begin{equation}\label{eq-nf}
   \mathbb{R}^{D}\ni \bz\longrightarrow (q_{pr}(\bz),\bm{\vphi}(\bz|\btheta))\in (\mathbb{R},\mathbb{R}^{D}),
\end{equation}
where the function $q_{pr}(\bz)$ is the probability density defining a {\em prior} distribution of random variable $\bz$. $\bvphi(\bz|\btheta)$ has to be a {\em bijection} which implies that if the input $\bz$ is drawn from $q_{pr}(\bz)$ then the output $\bphi$ is distributed according to 
\begin{equation}\label{eq-q-phi}
    q(\bphi|\btheta)= q_z(\bz|\btheta) \equiv  q_{pr}(\bz)\left|J(\bz|\btheta)^{-1}\right|,\quad \bphi=\bvphi(\bz|\btheta),
\end{equation}
where
\begin{equation}\label{eq_jac_def}
    J(\bz|\btheta)=\det \left(\diffp{\bvphi(\bz|\btheta)}{\bz}\right)
\end{equation}
is the determinant of the Jacobian of $\bvphi(\bz|\btheta)$. For practical reasons, normalizing flows are constructed in such a way that the Jacobian determinant is relatively easy to compute. 
The variational free energy $F_q$  defined in Eq.~\eqref{eq:shifted-DKL} can be rewritten in terms of $q_{pr}(\bz)$, $q_z(\bz|\btheta)$ and $\bvphi(\bz|\btheta)$ as
\begin{equation}
\label{eq. rep trick}
    F_q = \int\dl\bz\, q_{pr}(\bz) \left(\log q_z(\bz|\btheta)-\log P(\bvphi(\bz|\btheta))\right)=\E{\log q_z(\bz|\btheta)-\log P(\bvphi(\bz|\btheta))}_{q_{pr}(\bz)}.
\end{equation}
Eq.~\eqref{eq. rep trick} is known as the "reparametrization trick" because of the change of variables (reparameterization) from $\bphi$ to $\bz$ \cite{reparameterization}.
$F_q$ can be approximated as
\begin{equation}\label{eq:Fq-normalizing-flow}
      F_q \approx\frac{1}{N} \sum_{i=1}^N \left(\log q_z(\bz_i|\btheta)-\log P(\bvphi(\bz_i|\btheta))\right),\quad \bz_i\sim q_{pr}(\bz_i),
\end{equation}
where the $\sim$ symbol denotes that each $\bz_i$ is drawn from the distribution $q_{pr}(\bz_i)$.

\section{Gradient estimators}
\label{sec:grad-estimators}

The training of the machine learning model is done with the stochastic gradient descent (SGD) method and requires the calculation of the gradient of $F_q$ with respect to $\btheta$. The gradient is estimated based on a random, finite sample (batch) of $N$ configurations $\{\bphi\} = \{\bphi_1,\ldots,\bphi_N\}$.

\begin{figure}
\begin{center}
\begin{tikzpicture}[>={Stealth[length=2mm]}]
\newlength{\yoffset}
\setlength{\yoffset}{0cm}
\newlength{\lbloffset}
\setlength{\lbloffset}{3mm}
\tikzstyle{flow}=[double]

\begin{scope}[xshift = 4cm, yshift=\yoffset]
\node (z) at (0,0)  [rectangle,minimum size=5mm, align=center] {$\bz\sim q_{pr}(\bz)$};
\node (qz) at (0, 2)  [rectangle, minimum size=5mm] {$\bphi=\bvphi(\bz|\btheta), J(\bz|\btheta)$}; 
\node (qpt) [above of = qz, node distance=1cm] {$\qpt=q_{pr}(\bz)J(\bz|\btheta)^{-1}$};
\draw[->, densely dashed] (qz)--(qpt);
\draw  let \p1=(qz.west) in (\x1,-\yoffset-\lbloffset) node {(a)};
\draw[->, flow]   (z) -- (qz);
\end{scope}     

\begin{scope}[xshift=9.5cm, yshift=\yoffset]
\node (zp) at (-1cm,0)  [rectangle, minimum size=5mm] {$\bz'=\bvphi^{-1}(\bphi|\btheta), \bar J(\bphi|\btheta)$};
\node (phi) at (0.75cm,2)  [rectangle, minimum size=5mm] {$\bphi=\bvphi(\bz|\btheta)$};
\node (z) at (2.5,0) [rectangle, minimum size = 5mm] {$\bz\sim q_{pr}(\bz)$};
\node (qpt) [above of = zp, node distance=30mm] {$\qpt=q_{pr}(\bz')\bar J(\bphi|\btheta)$};
\draw[->] (zp)--(qpt);
\draw[->, flow]   (phi) -- (zp) ;
\draw[->, dashed, flow]   (z) -- (phi);
\draw  let \p1=(qpt.west) in (\x1-3mm,-\yoffset-\lbloffset) node {(b)};
\end{scope}    
\end{tikzpicture}
\end{center}
\caption{Schematic picture of two algorithms for gradient estimation discussed in the paper: a) reparametrization trick b) REINFORCE. Double line arrows represent the flow: upward-pointing arrows represent forward propagation, and downward-pointing arrows represent reversed propagation. Dashed arrows denote propagation which does not require gradient calculations.}
\label{fig_schem_alg}
\end{figure}
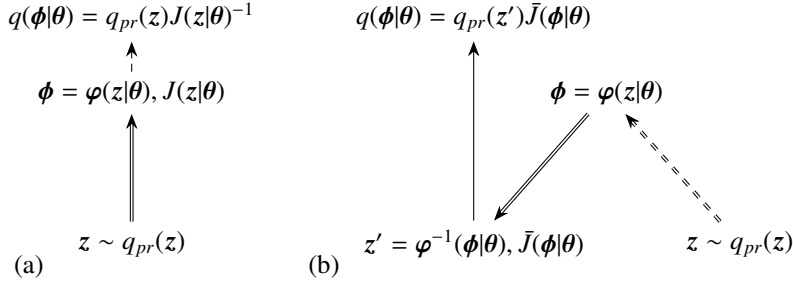

In the case of normalizing flows, this is pretty straightforward. We can directly differentiate the expression \eqref{eq:Fq-normalizing-flow} to obtain the gradient estimator $\g_{rt}[\{\bphi\}]$,
\begin{equation}\label{eq:g1}
   \begin{split}
        \diff{F_q}{\btheta} &\approx \g_{rt}[\{\bphi\}]
        \equiv\frac{1}{N}\sum_{i=1}^N\diff*{\left(\log q_z(\bz_i|\btheta)-\log P(\bvphi(\bz_i|\btheta))\right)}{\btheta},\quad \bz_i\sim q_{pr}(\bz_i).
   \end{split}
\end{equation}
This derivative can be calculated by popular machine learning packages like {\em e.g.} \pytorch\cite{PyTorch}  using automatic differentiation \cite{AD}.  All we have to do is to calculate \eqref{eq:Fq-normalizing-flow} and its gradient will be calculated in the backward propagation step. This is illustrated with the pseudo-code in Algorithm~\ref{alg:g2} (left) and shown schematically in Figure~\ref{fig_schem_alg}a. The striped-down version of the Python implementation is presented in Listing~\ref{lst:grt} in the \ref{app:implementation}.

While conceptually simple, this estimator has a considerable drawback as it requires calculating the gradient of the distribution $P(\bphi)$ with respect to the configuration $\bphi$,
\begin{equation*}
  \diff*{\log P(\bvphi(\bz_i|\btheta))}{\btheta} = \diffp*{\log P(\bphi)}{\bphi}[\bphi=\bvphi(\bz_i|\btheta)] \diffp{\bvphi(\bz_i|\btheta)}{\btheta}. 
\end{equation*}
In lattice field theories the probability $P$ is given by the {\em action} $S(\bphi)$,
\begin{equation}
    \log  P(\bphi(\bz|\btheta))=-S(\bphi(\bz|\btheta)) 
\end{equation}
and so calculating the gradient of $F_q$ requires the gradient of the action $S$ with  respect to the  fields $\bphi$. This may not pose large problems for {\em e.g.} $\phi^4$ theory discussed in \cite{bialas2022gradient} where the action is just a polynomial in $\bphi$. Other lattice field theories however, notably Quantum Chromodynamics with dynamical fermions, may have much more complicated actions including some representation of the non-local determinant of the fermionic matrix and the calculation of the action gradient may be impractical.

The free energy $F_q$ can also be approximated without any reparameterization
\begin{equation}
    F_q\approx \frac{1}{N}\sum_{i=1}^N \left(\log q(\bphi_i|\btheta) - \log P(\bphi_i)\right),
    \quad \bphi_i \sim q(\bphi_i|\btheta)
\end{equation}
Unfortunately, this expression cannot be differentiated directly with respect to $\btheta$ as the distribution from which $\bphi_i$ are drawn depends on $\btheta$. Instead, the
REINFORCE algorithm relies on first differentiating the exact formula \eqref{eq:shifted-DKL} \cite{VANPRL, Rezende2016}
\begin{equation}
\begin{split}
\label{eq:Fq-grad-autoregessive}
 \diff{F_q}{\btheta} & = \int\dl\bphi \, \diffp{q(\bphi|\btheta)}{\btheta} \left(\log q(\bphi|\theta)-\log P(\bphi)\right) \\
  &\phantom{=}+\int\dl\bphi \,  q(\bphi|\btheta) \diffp*{\log q(\bphi|\theta)}{\btheta}.
  \end{split}
\end{equation}
The last term in the above expression is zero because it can be rewritten as the derivative of a constant,
\begin{equation}\label{eq-const-der}
  \E{\diffp{\log q(\bphi|\theta)}{\btheta}}_{\qpt} = \int\dl\bphi \,  \diffp{ q(\bphi|\theta)}{\btheta} =\diffp*{ \underbrace{\int\dl\bphi \,   q(\bphi|\theta)}_1}{\btheta}  = 0.
\end{equation}
The first term in the expression \eqref{eq:Fq-grad-autoregessive} can be further rewritten as
\begin{equation}\label{eq-KL-grad-simpl}
\begin{split}
     \diff{F_q}{\btheta} 
        & =\int\dl\bphi \, q(\bphi|\btheta)\diffp{\log q(\bphi|\btheta)}{\btheta} \left(\log q(\bphi|\theta)-\log P(\bphi)\right)\\
        &=\E{\diffp{\log q(\bphi|\btheta)}{\btheta}\left(\log q(\bphi|\theta)-\log P(\bphi)\right)}_{q(\bphi|\btheta)}
     \end{split},
\end{equation}
and approximated as
\begin{equation}\label{eq-KL-grad-approx}
   \diff{F_q}{\btheta}\approx \g_{\widetilde{RE}}[\{\bphi\}]\equiv\frac{1}{N} \sum_{i=1}^N \diffp{ \log q(\bphi_i|\btheta)}{\btheta} \left(\log q(\bphi_i|\btheta)-\log P(\bphi_i)\right),
\end{equation}
which defines another gradient estimator $\g_{\widetilde{RE}}[\{\bphi\}]$. Contrary to $\grt$, this estimator does not require calculating the derivatives of $P(\bphi)$.

In practice, this estimator has a huge variance and one has to use some variance-reducing method \cite{VANPRL,bialas2022gradient, pathGradient,Rezende2016}. Following the Ref.~\cite{VANPRL} we define the final  version of this estimator as  
\begin{equation}\label{eq:def-g2}
   \gri[\{\bphi\}]= 
   \frac{1}{N}\sum_{i=1}^N 
   \diffp{ \log q(\bphi_i|\btheta)}{\btheta}  \left(s(\bphi_i|\btheta)-\overline{s(\bphi|\btheta)_N} \right),
\end{equation}
where 
\begin{equation}\label{eq:signal}
     s(\bphi|\btheta) \equiv  \log q(\bphi|\btheta)-\log P(\bphi)\quad\text{and}\quad  \overline{s(\bphi|\btheta)_N} =\frac{1}{N}\sum_{i=1}^{N} s(\bphi_i|\btheta).
\end{equation}

Contrary to $\g_{rt}$ the $\gri$ estimator is slightly  biased
\begin{equation}\label{eq:g2-bias}
\E{\gri[\{\bphi\}]}=\frac{N-1}{N}\E{\grt[\{\bphi\}]}.
\end{equation}
The proof of this fact is presented in \cite{bialas2022gradient}. Of course, such multiplicative bias does not play any role when the estimator is used in the gradient descent algorithm and is very small anyway when $N\sim 10^3$. For all practical purposes, we can treat both estimators as unbiased, so any differences must stem from higher moments, most importantly from the variance. 
Although not much can be said about the variances of these estimators in general, we can show that for perfectly trained model {\em i.e.} when $\qpt=p(\bphi)$
\begin{equation}
\var{\gri[\{\bphi\}] }_{\qpt=p(\bphi)}=0
\end{equation}
The proof is presented in  \cite{bialas2022gradient}. 
As for the estimator $\g_{rt}$, we cannot make any claims as to the value of its variance but in Ref.~\cite{bialas2022gradient} we showed that it does not need to vanish even for $\qpt=p(\bphi)$.


Expression \eqref{eq:def-g2} is not yet ready for direct calculation in the code. The $\gri$ estimator requires the $\qpt$ function, and while it is not explicit in the normalizing flows formulation \eqref{eq-nf}, it can be inferred from 
Eq.~\eqref{eq-q-phi}. 
Using the fact that the Jacobian determinant of transformation $\bvphi^{-1}(\bphi|\btheta)$ 
\begin{equation*}
 \bar J(\bphi|\btheta)\equiv \det \left(\diffp{\bvphi^{-1}(\bphi|\btheta)}{\bphi}\right)
\end{equation*}
is the inverse of Jacobian determinant of $\bvphi(\bz|\btheta)$, 
\[
\bar J(\bphi|\btheta)= J(\bz|\btheta)^{-1},
\]
we can write $\qpt$ as
\begin{equation}\label{eq:q-phi-inv}
\qpt=q_{pr}(\bz')\left|\bar J(\bphi|\btheta)\right|\quad \bz'=\bvphi^{-1}(\bphi|\btheta).
\end{equation}
The final expression for the \reinf{} gradient estimator is

\begin{equation}\label{eq:def-g2-normalizing}
   \gri[\{\bphi\}]= 
   \frac{1}{N}\sum_{i=1}^N  \left(s(\bphi_i|\btheta)-\overline{s(\bphi|\btheta)_N} \right)
   \diff*{\log\left( q_{pr}\left(\bvphi^{-1}(\bphi_i|\btheta)\right)\left|\bar J(\bphi_i|\btheta)\right|\right)}{\btheta} ,\quad \bphi_i=\bvphi(\bz_i),\; \bz_i\sim q_{pr}(\bz_i).
\end{equation}
The way to calculate this expression using the auto-differentiation is to calculate a {\em pseudo-loss} function in which parts of the expression will be "protected" from gradient calculations as marked below  

\begin{equation}\label{eq:def-g2-normalizing-loss}
   L_{RE}[\{\bphi\}]= 
   \frac{1}{N}\sum_{i=1}^N  
   \underbrace{\left(s(\bphi_i|\btheta)-\overline{s(\bphi|\btheta)_N} \right)}_{\text{ no grad.}}
   \log\left( q_{pr}\left(\bvphi^{-1}(\bphi_i|\btheta)\right)\left|\bar J(\bphi_i|\btheta)\right|\right),\quad \underbrace{\bphi_i=\bvphi(\bz_i),\; \bz_i\sim q_{pr}(\bz_i)}_{\text{no grad.}}.
\end{equation}
and then using the backpropagation to calculate the gradient.  Please note that $L_{RE}$ it is not a real loss function, as it is not minimized by the SGD (only part of it's expression is differentiated). The real loss function remains the $F_q$. $L_{RE}$  is just an auxiliary expression used to calculate the gradient of the $D_{KL}$.   
This illustrated with the pseudo-code in Algorithm~\ref{alg:g2} (right) and schematically in Figure~\ref{fig_schem_alg}b. This requires running the flow two times: forward to obtain $\bphi_i$, then backward to calculate $\bz'$, but the gradients have to be calculated only on the second pass. The stripped-down version of the actual \texttt{Python} code is presented in Listing~\ref{lst:gre} in the \ref{app:implementation}.

\begin{algorithm}
\caption{\label{alg:g2}Calculation of the $\grt$ (left) and $\gri$ (right) estimators for normalizing flows. In each case the estimator will be calculated during the  backpropagation step on the $loss$.  The hash symbol denotes comments.} 
\hfill\begin{minipage}[t]{\textwidth/2}
\begin{algorithmic}[1]
\State\Comment{generate $\bphi$}
\State Switch on gradient calculations
\State $\bz\sim q_{pr}(\bz)$ \Comment{ generate $z$ from prior distribution}
\State $\bphi \gets \bvphi(\bz|\btheta)$ \Comment{Forward pass}
\State\Comment{Calculate $q(\bphi|\btheta)$}
\State $q\gets q_{pr}(\bz)\det \left(\diffp{\bvphi^{-1}(\bphi|\btheta)}{\bphi}\right)$ 
\State\Comment{Calculate loss}
\State $loss \gets \log q(\bphi|\btheta)-\log P(\bphi)$
\end{algorithmic}
\end{minipage}\begin{minipage}[t]{\textwidth/2}
\begin{algorithmic}[1]
\State\Comment{generate $\bphi$}
\State Switch off gradient calculations
\State $\bz\sim q_{pr}(\bz)$ \Comment{ generate $z$ from prior distribution}
\State $\bphi \gets \bvphi(\bz|\btheta)$ \Comment{Forward pass}
\State\Comment{Calculate signal}
\State $s \gets \log q(\bphi|\btheta)-\log P(\bphi)$
\State\Comment{Calculate $q(\bphi|\btheta)$}
\State Switch on gradient calculations
\State $\bz'\gets \bvphi^{-1}(\bphi|\btheta)$ \Comment{Backward pass}
\State $q\gets q_{pr}(\bz'|\btheta)\det \left(\diffp{\bvphi^{-1}(\bphi|\btheta)}{\bphi}\right)$ 
\State\Comment{Calculate loss}
\State $loss \gets \log q \times (s-\bar{s})$
\end{algorithmic}
\end{minipage}\hfill
\end{algorithm}

\section{Schwinger model}
\label{sec:schwinger}

We compare the \reinf{} and \rt{} estimators simulating  the two-dimensional Schwinger model with two flavours of Wilson fermions defined on a $L\times L$ lattice. The action consists of two parts: the pure gauge plaquette action and the fermionic determinant which following Ref.~\cite{PhialaSchwinger} we calculate directly (using built-in \pytorch function). 
\begin{equation}
    S(U) = -\beta\sum_{x}\re P(x) -\log\det D[U]^\dagger D[U]
\end{equation}
where $P(x)$ is the plaquette 
\begin{equation}
    P(x)=U_1(x) U_0(x+\hat{1}) U_1^\dagger(x+\hat{0}) U_0^\dagger (x).
\end{equation}
$U_\mu(x)$ is a link starting from $x$ in direction $\mu=0,1$ and $\hat\mu$ is the displacement vector of one lattice site in the direction $\mu$.
The Wilson-Dirac operator is defined as
\begin{equation}
    D[U](y,x)^{\alpha\beta}=\delta(y-x)\delta^{\alpha\beta}
    -\kappa \sum_{\mu=0,1}\left\{[1-\sigma^\mu]^{\beta\alpha}U_\mu(y-x+\hat\mu)\delta(y-x+\hat\mu)+
    [1+\sigma^\mu]^{\beta\alpha}U^\dagger_\mu(y-\hat\mu)\delta(y-x-\hat\mu)
    \right\}
\end{equation}
where $\sigma^\mu$ are the Pauli matrices. 

In our implementation, we have recreated the normalizing flow architecture described in  Refs.~\cite{PhialaSchwinger, PhialaEquivariant}. We did this on top of the code provided in \cite{albergo2021introduction}, which provided the implementation of the pure $U(1)$ gauge model with non-compact projection as the plaquette coupling layer. We have provided our own implementation of the circular splines plaquette coupling layer \cite{NeuralSplines,ToriSphere}, more complicated masking patterns with $2\times 1$ loops described in \cite{PhialaSchwinger} and fermionic action of the Schwinger model. More details can be found in \ref{app:implementation}.
  The full code is provided in \cite{code}, which is, up to our knowledge, the only open access code for normalizing flow sampling of the Schwinger 2D model.

Following \cite{PhialaSchwinger}, we concentrate in this paper only on a single point in the phase space: $\beta=2$ and $\kappa=0.276$ where the model is expected to be at criticality, as this is where the most severe critical slowing down is expected. 

\section{Results}
\label{sec:results}

We start with $16\times 16$ lattice for direct comparison with \cite{PhialaSchwinger}. 
To monitor the progress of training we have used the effective sample size (ESS),
\begin{equation}\label{eq:ess-def}
\ess[\{{\bphi}\}] = \frac{\E{w(\bphi)}^2_{q(\bphi|\btheta)}}
{\E{w(\bphi)^2}_{q(\bphi|\btheta)}}
\approx\frac{\left(\sum_{i=1}^N w(\bphi_i)\right)^2}{N\sum_{i=1}^N w(\bphi_i)^2},\qquad \bphi_i\sim q(\bphi_i|\btheta),
\end{equation}    
where 
\begin{equation}
    w(\bphi) =\frac{P(\bphi)}{q(\bphi|\btheta)} 
\end{equation}
are so called unnormalized importance ratios. It is easy to see that $0\le\ess\le1$ and $\ess=1$ if and only if $q(\bphi|\btheta)=p(\bphi)$, namely for a perfectly trained network. 

\begin{figure}
    \centering
    \includegraphics[width=0.475\textwidth]{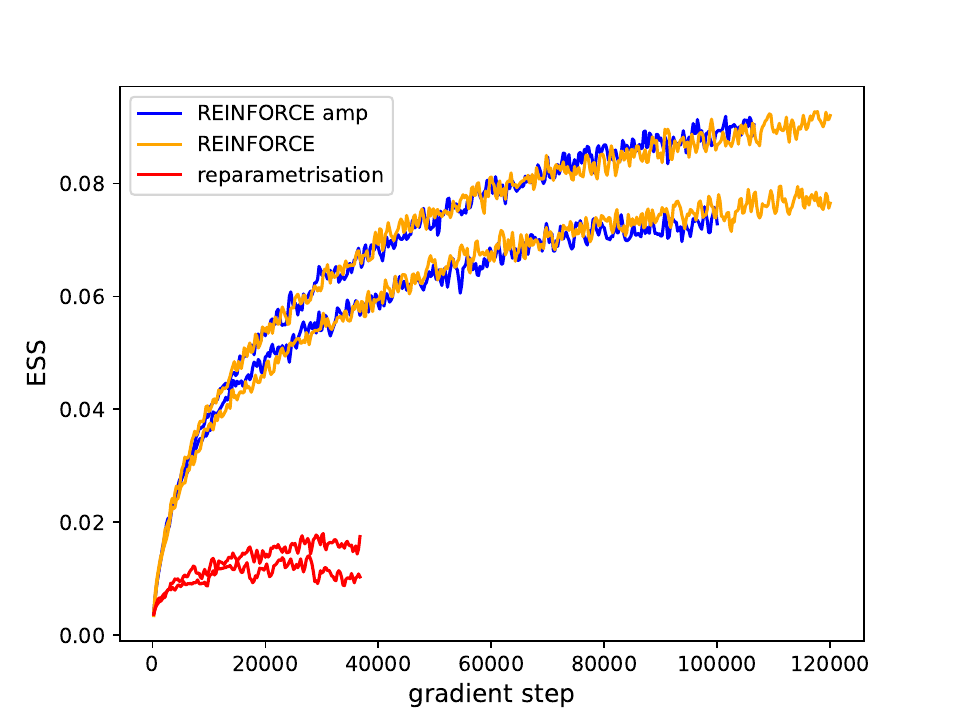}\includegraphics[width=0.475\textwidth]{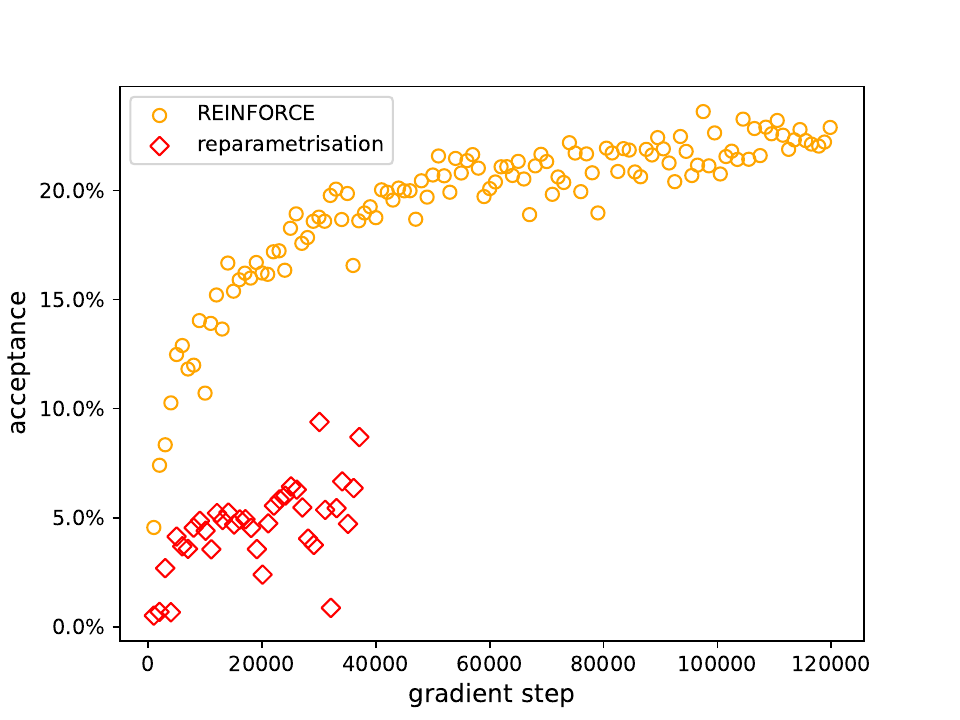}
    \caption{Training history for the Schwinger model on a  $16\times 16$ lattice at criticality $\beta=2.0$, $\kappa=0.276$. Each gradient step was calculated on a batch of $3\times512$ samples (the batch was split into three parts to fit on the GPU). Left: the effective sample size (ESS) defined in eq.~(\ref{eq:ess-def}) as a function of the number of gradient steps for two gradient estimators. Red curves were obtained using REINFORCE estimator with automatic mixed precision (\texttt{amp}) which enabled the use of tensor cores on GPU using half-float precision. We present the history of two different runs for each estimator. Right: the acceptance rate of the MCMC algorithm calculated for a particular state of the network after a given number of gradient steps.}
    \label{fig:ess_schwinger-L16}
\end{figure}

The results are presented in the left panel of Figure \ref{fig:ess_schwinger-L16}.   The ESS was calculated after each gradient step on a small batch of $1536$ configurations. Those values fluctuate wildly for the batch size we used. In order to obtain smoother curves we present an average of 500 consecutive measurements. We made several runs for each estimator and chose to present two of them as most typical. 

In the right panel in Figure \ref{fig:ess_schwinger-L16} we show the acceptance of the Metropolis-Hastings step \eqref{eq:accept-reject}. This was calculated offline: for a given state of the neural network, we generated a Markov Chain of $2^{16}$ configurations and measured the acceptance rate. Measurements were repeated after 1000 gradient steps in the training process. We show only the results of the run with the best ESS.  

The first thing to notice is that the REINFORCE estimator leads to much more efficient training: the ESS and acceptance rate grow much faster with gradient steps than for the r.t. estimator. The ESS reached after 120k gradient steps is 4-5 times larger for RE than in the r.t. case. The training using r.t. was finished at 40k gradient steps, because no further improvement was observed during the training.

The superiority of the RE in terms of the training efficiency was already observed in \cite{bialas2022gradient} in the case of the $\phi^4$ theory and was attributed to the smaller variance of this estimator. Please note, however, that the training efficiency may depend on the actual physical parameters of the model and may vary. Another possible cause is the numerical instability of the \rt{} estimator. In Ref.~\cite{PhialaSchwinger} the double precision was used for all but the neural network evaluations. As our aim was not to reproduce exactly the results of that paper, but to compare two different gradient estimators, we did all our calculations in single precision. We did not encounter any problems while training using the REINFORCE estimator. We also tried the automatic mixed precision (\texttt{amp}) features of the PyTorch library which enabled the use of tensor cores on GPU using half-float precision. This was not possible for $\grt$ as it crashed almost immediately. The crashes also happened when using \texttt{amp} with $\gri$ estimator but very rarely. Nevertheless, because of the possibility of encountering crashes and the fact that the speed up was only around $20\%$ (see table~\ref{tab:timings}) it is probably not worthwhile to use \texttt{amp} in this situation. 

\begin{figure}
    \centering
    \includegraphics[width=0.9\textwidth]{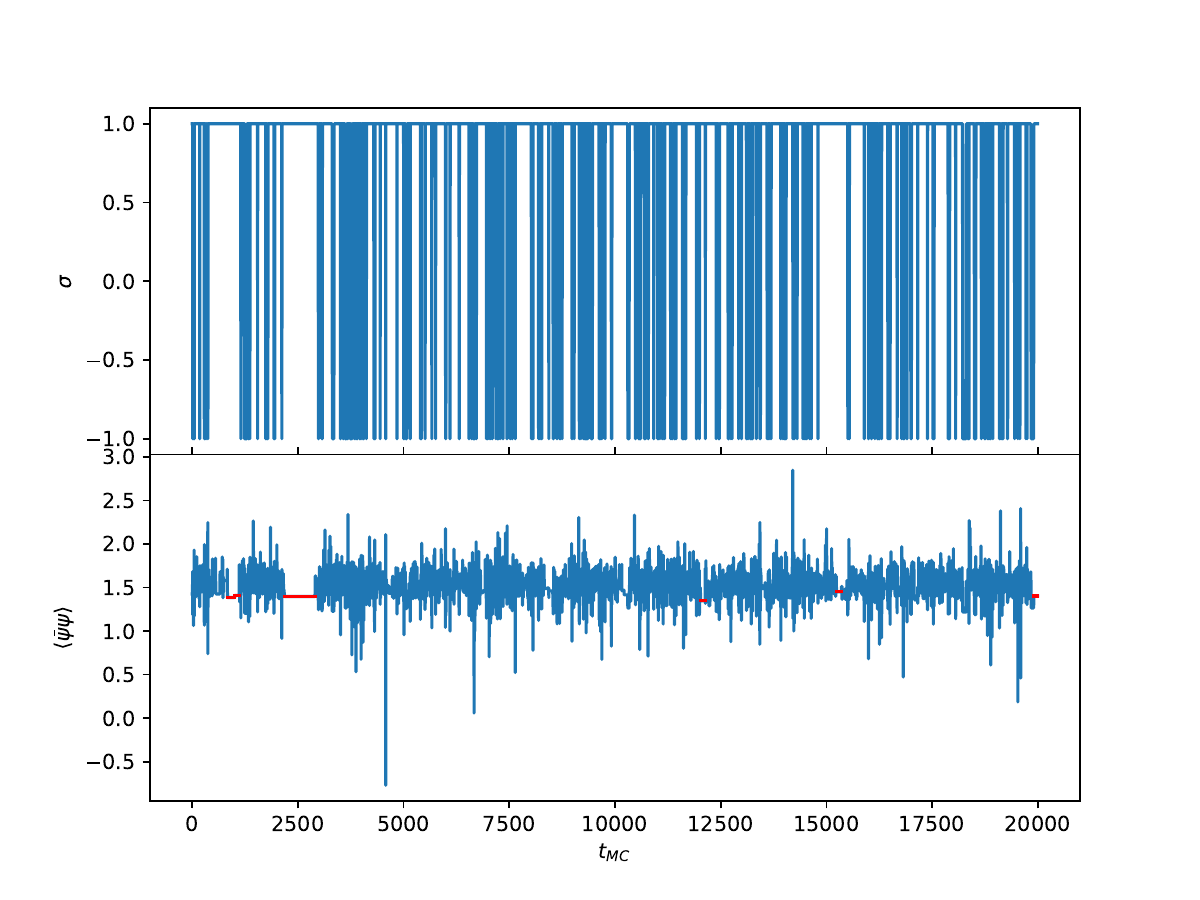}
    \caption{An excerpt from the Monte Carlo history for $\sigma$ (upper panel) and chiral condensate (lower) for $L=16$. Series of non-accepted configurations (bridges) that are over 100 in length are marked in red.}
    \label{fig:tunelling-L16}
\end{figure}

After $\sim120k$ gradients updates we achieved the acceptance rate of $22\%$ and autocorrelation time of approximately nine Monte Carlo steps for the chiral condensate,
\begin{equation}\label{eq:cond}
   \avg{ \bar\psi \psi } =\frac{1}{V\tr D^{-1}[U]}.
\end{equation}
In figure~\ref{fig:tunelling-L16} we present an excerpt from the Monte Carlo history for the chiral condensate and 
\begin{equation}
    \sigma = \sign (\re \det D).
\end{equation}
The value of $\sigma$ is positive (negative) for even (odd) topological sectors, and the changes in its value are correlated with tunneling events \cite{PhialaSchwinger}. We include this plot to show that the algorithm is not "frozen" in one of the topology sectors. 
We can see some small   "bridges" characteristic for the metropolized independent sampler with low acceptance, when the algorithm is stuck at a single configuration and new proposals get rejected. These bridges are responsible for the autocorrelation time. 

\begin{figure}
    \centering
    \includegraphics[width=0.475\textwidth]{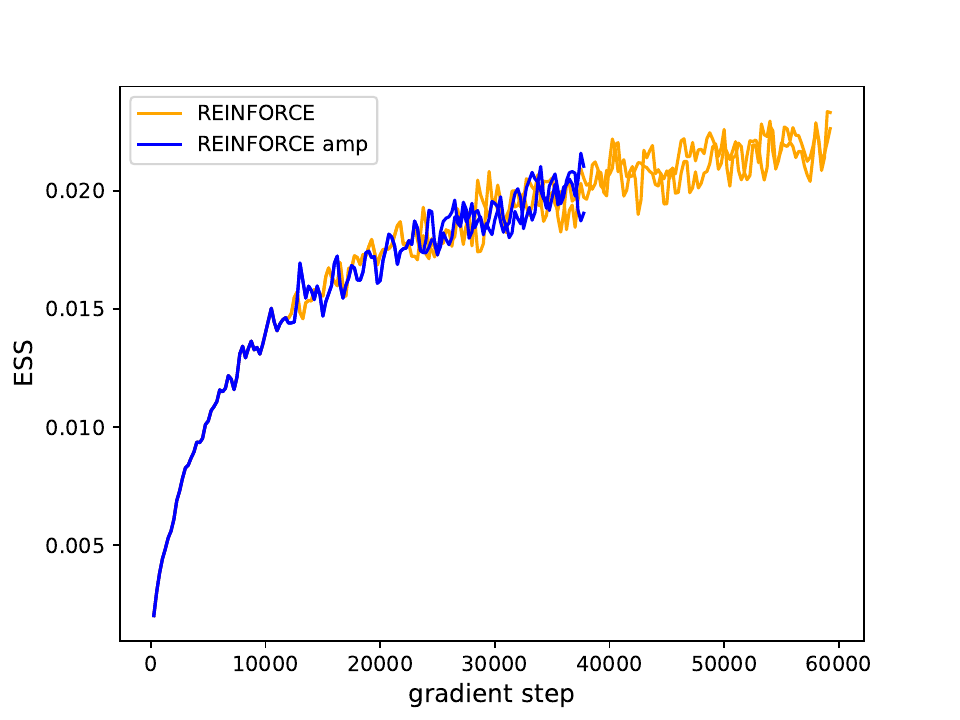}\includegraphics[width=0.475\textwidth]{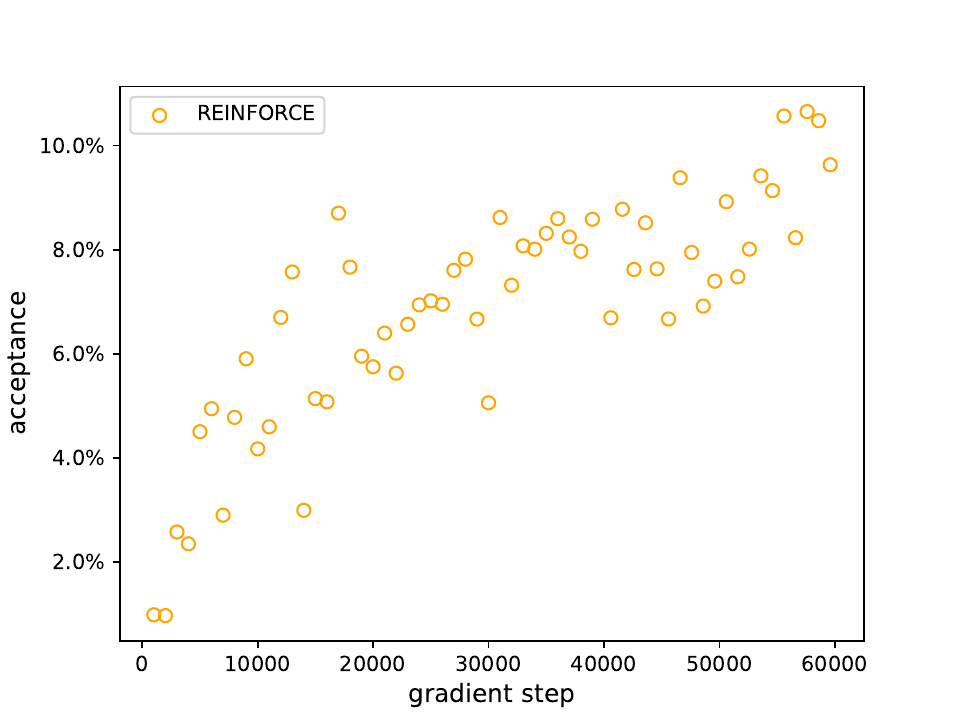}
        \caption{Training history for the Schwinger model on a $24\times 24$ lattice at criticality $\beta=2.0$, $\kappa=0.276$. Each gradient step was calculated on a batch of $4\times384$ samples (the batch was split into four parts to fit on the GPU). Left: the effective sample size (ESS) defined in eq.~(\ref{eq:ess-def}) as a function of the number of gradient steps for the REINFORCE gradient estimator. Red curves were obtained using REINFORCE estimator with automatic mixed precision (\texttt{amp}) which enabled the use of tensor cores on GPU using half-float precision. We present the history of two different runs for each estimator. Right: the acceptance rate of the MCMC algorithm calculated for a particular state of the network after a given number of gradient steps.}
    \label{fig:ess_schwinger-L24}
\end{figure}

\begin{figure}
    \centering
    \includegraphics[width=0.9\textwidth]{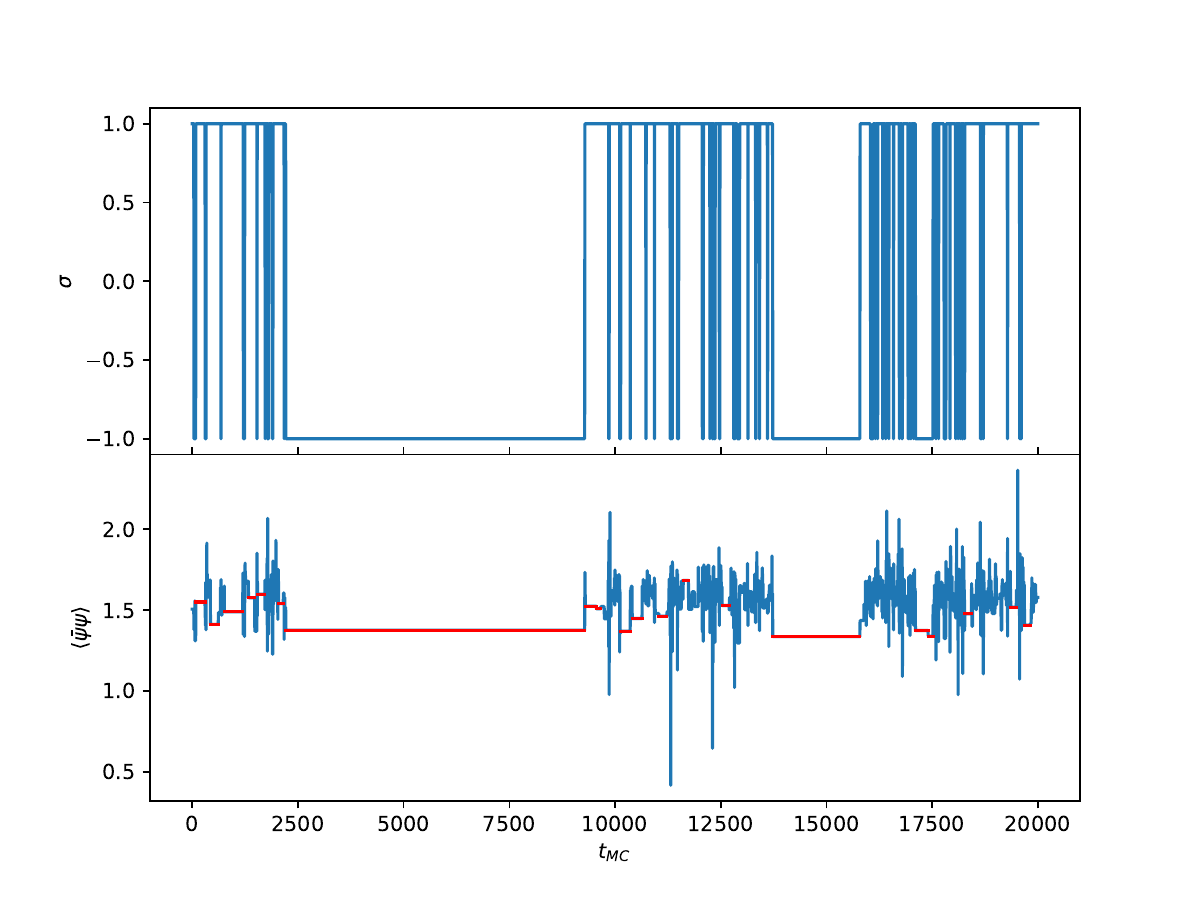}
    \caption{An excerpt from the Monte-Carlo history for $\sigma$ and chiral condensate for $L=24$. Series of non-accepted configurations (bridges) that are over 100 in length are marked in red.}
    \label{fig:tunelling-L24}
\end{figure}

Due to the improvements presented above, the REINFORCE estimator allows to simulate bigger systems and we have also tried the $24\times 24$ lattice.  We present the results of ESS and acceptance rate as a function of gradient steps in figure~\ref{fig:ess_schwinger-L24}. We do not include the $\grt$ in the comparison considering its poor performance which would be only exacerbated on bigger lattices. 
An excerpt from the Monte Carlo history for $L=24$ is presented in figure~\ref{fig:tunelling-L24}. Bridges are much longer than for $L=16$ as the acceptance is smaller $\sim8\%$ and, in consequence, the autocorrelation time is much larger $\sim68$, but still, the algorithm is able to explore many topological sectors.

The significant advantage of the \reinf{} estimator over \rt{} visible in figure~\ref{fig:ess_schwinger-L16} can depend on the parameters of the Schwinger model ($\beta$ and $\kappa$) as well as the hyper-parameters used in training. As this contribution is focused on purely computational advantages of the \reinf{} estimator, we have not explored this issue in depth. We have only made several short runs with different values of $\kappa$ both below and above the critical point. Please note that we have not attempted to optimize the hyper-parameters for each $\kappa$ separately. The results are presented in the Table~\ref{tab:kappa-convergence}. The \reinf{} estimator performs best away from the critical point (attains the higher values of ESS) which is not surprising. Up to $\kappa=0.270$ it maintains a modest advantage over the \rt{} estimator. However, as we approach the critical point the performance of the $\rt{}$ estimator degrades dramatically and the training effectively stops beyond the critical point, while the training with \reinf{} estimator proceeds without problems, at least as it concerns the values of ESS. These observations merit further in-depth study which is beyond the scope of this contribution. 

\begin{table}
    \centering
    \begin{tabular}{||c|rrrrrr||}\hline\hline
            & \multicolumn{6}{c||}{$\kappa$}\\\hline
         & 0.200 & 0.250 &0.270 & 0.276& 0.300 & 0.350\\\hline
        \reinf{} & 0.90& 0.81 & 0.26& 0.06& 0.29 & 0.59\\\hline
        \rt{} & 0.79& 0.71& 0.21& 0.01& 0.00 & n/a\\\hline\hline
    \end{tabular}
    \caption{ESS obtained after training for  $2\times10^5$ gradient updates for different values of $\kappa$ and $\beta=2.0$. Shown is the average value of the last $500$ updates. Non-available (n/a) value indicates that this run was not completed due to numerical instabilities. }
    \label{tab:kappa-convergence}
\end{table}

In the next section we present the detailed timing measurements for single gradient update for each estimator. Contrary to the convergence rate, these should not depend on $\beta$  or $\kappa$.

\subsection{Timing}

Due to the fact that the \reinf{}  algorithm does not require propagating the gradient through the complicated determinant, we expect it to also reduce the time required for the evaluation of one gradient step. To check this, we compared the wall clock time of 100 gradient steps during the training for \rt{} and \reinf{}. The times of the latter estimator were measured separately for \texttt{amp} features on and off. The batch size was 1536, but it was split differently depending on the lattice size to fit on a single GPU.

Results are shown in table~\ref{tab:timings} for different lattice sizes from $L=8$ to $L=24$. We see that the REINFORCE algorithm starts to outperform the reparameterization trick at $L=12$. At $L=24$ one gradient step with the RE estimator is almost nine times faster than with the r.t.. Activating the \texttt{amp} features we gain an additional $15-20\%$ speed-up for \reinf{}. In the case of r.t., \texttt{amp} cannot be used due to numerical instability.

\begin{table}
    \centering
    \begin{tabular}{|rrr|r|rr|rr|}
    \hline\hline
        & & & \multicolumn{1}{c|}{r.t.} & \multicolumn{2}{c|}{\reinf{} } & \multicolumn{2}{c|}{\texttt{amp} }\\\hline
        L  &bs. & nb. & t[s] &t[s]& sp. & t[s]& sp. \\\hline\hline
     8 & 1536 & 1 & 113  & 137& 0.82 & 120& 0.94\\\hline
    12 & 1536 & 1 & 242  & 189& 1.28 &159&1.52 \\\hline
    16 & 768 & 2 & 784  & 331&  2.37 &285& 2.75\\
    16 & 1536 & 1 &   & 268&  $\underbar{2.93}$ &215& $\underbar{3.65}$\\\hline
    20 & 512 & 3 & 2411   &527&  4.50 & 460& 5.24\\
    20 & 768 & 2 &    &450&  $\underbar{5.36}$  & 372& $\underbar{6.48}$ \\\hline
    24 & 384 & 4 & 6496 &786& 8.26 & 678& 9.58\\
    24 & 512 & 3 &  &690& $\underbar{9.41}$  & 598& $\underbar{10.86}$ \\\hline
    \end{tabular}
    \caption{Time (t) in seconds required for 100 gradients steps.  Missing entries in the table could not be computed because these $L$ and batch size combinations did not fit into GPU memory.  "bs." stands for batch size as sent to GPU, "nb." stands for the number of such batches used to calculate the gradient and "sp." is the speed-up factor compared to the \rt{} estimator. When a \rt{} entry is missing the speed-up factor was calculated relative to the best \rt{} time on the same lattice size, those sp. factors are underlined. \texttt{amp} denotes the use of automatic mixed precision entailing the use of tensor cores. Timings were measured on NVIDIA A100-SXM4-40GB GPU. }
    \label{tab:timings}
\end{table}

\begin{table}
    \centering
    \begin{tabular}{|rr|rrr|rrr|} 
    \hline\hline
    L & b.s.& \multicolumn{3}{c|}{r.t.}  & \multicolumn{3}{c|}{\reinf{} }  \\\hline
      & & total & loss & back. & total & loss & back. \\\hline\hline
        8   &  1536 &   0.97& 0.34 & 0.63   &1.20   & 0.61 & 0.59  \\\hline
        12  &  1536 &   2.25& 0.48 & 1.76   &1.70   & 0.81 & 0.89  \\\hline
        16  &  768  &   3.85& 0.53 & 3.32   &1.57   & 0.82 & 0.75  \\
        16  &  1536 &   & &                 &2.48   &  1.14 & 1.35\\\hline
        20  &  512  &   7.98& 0.67 & 7.03   &1.69   & 0.93&  0.75 \\
        20  &  768  &       &   &           &2.14   & 1.10&  1.05 \\\hline
        24  &  384  & 16.27& 0.84 & 15.43   & 1.85  & 1.08&  0.77 \\
        24  &  512  & &  &                  &2.23   & 1.24&  0.99 \\
         \hline\hline
    \end{tabular}
    \caption{Timings in seconds for one call to loss function and backward propagation. 
    "loss" column presents timings for one call to the loss function \texttt{grt\_loss} or  \texttt{gre\_loss} (see listings~\ref{lst:grt} and \ref{lst:gre} ). "back." presents timings to one call to \texttt{.backward()} method. 
    Timings were measured on NVIDIA A100-SXM4-40GB GPU. Missing entries could not be computed because of the memory constraints. 
}
    \label{tab:detailed_timings}
\end{table}

To see why \reinf{}  is faster than r.t. we have performed detailed timing measurements on a single call to loss function and subsequent backward propagation using CUDA event timers. The results are summarized in table~\ref{tab:detailed_timings}. The  "loss" column presents timings for one call to the loss function \lstinline!grt_loss! or \lstinline!gre_loss! (see listings~\ref{lst:grt} and \ref{lst:gre} ). The "back." columns present timings of one call to the \lstinline!.backward()! method. 

We see that \reinf{}  is slightly slower than r.t. in calculating the loss (which is understandable since it requires an additional pass through the network). 
On the other hand, the backward propagation part is several times faster in \reinf{}  For the r.t. case, the backward propagation is the bottleneck of the algorithm, especially at larger lattice sizes. In the case of RE, the timings of the loss computation and backward propagation are comparable.

The difference in backward propagation timing between the two gradient estimators can be attributed to the size of the computational graph constructed during the forward pass. When calculating any expression that involves tensors requiring gradient computations, \pytorch during the forward pass constructs a directed acyclic graph  (DAG) containing the information needed to calculate gradients during the subsequent backward pass \cite{autograd}. The DAG stores not only the operations needed to be performed but also all partial results (tensors) when needed (see next section). Thus, each operation (function) in \pytorch has a forward and backward implementation. Forward implementation evaluates this function and optionally registers in the DAG a node with backward implementation, storing intermediate results if necessary. Backward implementation calculates the gradient of this function with respect to its inputs using stored intermediate results. For example, if calculating $\bm{x}^2$ \pytorch would add a node to the graph calculating $2\cdot\bm{x}$ and store $\bm{x}$ in this node. 

We have counted the number of nodes in  DAG for different values of $L$ and the results are presented in table~\ref{tab:tree-size}. As we can see in the case of the REINFORCE algorithm, the number of nodes is constant. This is because all operations relevant to the gradient computation were implemented as calls to \pytorch functions without any loops depending on the size of the lattice. The architecture of the neural networks was also not changed. This resulted in exactly the same graph with the only difference being that each node of the graph would process bigger tensors in case of bigger lattices or bigger batches.  
In the case of the reparameterization trick, we can see that the size of the tree grows with the lattice size. In fact, the number of nodes is {\em exactly} $36L^2+12036$. This $\propto L^2$ dependence comes from the gradient calculations with the fermionic determinant. While this is also a single call to a \pytorch function, it first requires an assembly of the Dirac matrix, which is done using \texttt{for} loops. Each iteration of the loops adds its operation to the DAG  resulting in the growing size.  While the size of the Dirac operator is $\propto L^4$, it has only $\propto L^2$ non-zero elements thus explaining the scaling of the tree growth.  

Using the \pytorch feature allowing to register callbacks on the nodes of the computational graph we have measured the time taken to execute each individual node. The results are presented in the table~\ref{tab:node-timings} where we list the top 10 time-consuming functions for each algorithm. For each function we present the number of times this function was called (num. op.) {\it i.e.} the number of nodes in the DAG performing this function and the total time (in milliseconds)  taken by all these calls together. The functions presented in this table are \pytorch internal (backward) functions implementing gradient computations. One can deduce which function gradient was calculated by omitting the \texttt{Backward0} suffix. We have found out that most of the time difference between the two algorithms can be attributed to the \texttt{CopySlices} function which takes over 20 times more time in r.t. than in the RE case. The \texttt{LinalgSlogdetBackward0} function responsible for determinant calculation has a negligible effect. This function is not present at all in \reinf{}, as this estimator does not require backpropagating through the determinant. The \texttt{CopySlices} function is the only function in the table that does not have the \text{Backward0} suffix.  That, and the name, would again point out to the assembly of the Dirac matrix being the most time-consuming operation. Please note that those measurements were taken on a different GPU so they cannot be directly compared to the results from Tables \ref{tab:timings} and \ref{tab:detailed_timings}. This breakdown is of course characteristic of only this particular model but we are of the opinion that such analysis may be beneficial in general. The code for obtaining such measurements is also provided in \cite{code}.  

\begin{table}
    \centering
\begin{tabular}{|r|rrrrrrrr|}\hline\hline
L & 4 &8 & 12 & 16 & 20 & 24 & 28 & 32\\\hline\hline
   r.t. & 12612 & 14340 & 17220&  21252&  26436 & 32772 & 40260 & 48900 \\\hline
   \reinf{}  & 12989 & 12989  & 12989& 12989 & 12989 & 12989 & 12989 &  12989\\\hline\hline
\end{tabular}
\caption{The number of nodes in the computational graph for different loss functions. The \reinf{} loss function builds a tree which size is independent of the lattice size, while the size of the tree for the \rt{} loss function is {\em exactly} $36L^2+12036$. }
\label{tab:tree-size}
\end{table}

\begin{table}
    \centering
    \begin{tabular}{|rr>{\tt}l|rr>{\tt}l|}
    \hline\hline
    \multicolumn{3}{|c|}{r.t} & \multicolumn{3}{c|}{REINF.}\\ \hline
    time[ms] & num. op.  & \textrm{name} & time[ms] & num. op.  & {\rm name} \\\hline
           2743.56 & 2027 &CopySlices  &  270.27  & 1342  & IndexBackward0   \\      
    189.55  & 960 & IndexBackward0     & 47.78  & 144 & ConvolutionBackward0  \\
     146.63  & 144 & ConvolutionBackward0 & 129.15 &  1003 & CopySlices \\
     110.57 & 3406 & SliceBackward0     &  105.70 &  2378 &SliceBackward0 \\
      82.90  &   1 & LinalgSlogdetBackward0 & 42.19 &  1813 & MulBackward0  \\
      48.68 & 3677 & MulBackward0        &  18.39 & 1475 &SubBackward0  \\
      25.67  &3925 & SelectBackward0     &  11.70 &   96 & LeakyReluBackward0  \\
      11.72  &  96 & LeakyReluBackward0  & 8.55  & 848 & SelectBackward0  \\
      10.47 & 1002 & SubBackward0        & 6.60  & 144 & DivBackward0\\
      7.97  & 192 & DivBackward0 &  5.60  & 283 &WhereBackward0\\\hline\hline
    \end{tabular}
    \caption{Top ten operations, in terms of time used, performed while backpropagating through the computational graph for a $16\times 16$ lattice and batch size of $512$. The table presents the number of times this function was called (num. op.) and the total time (in milliseconds) taken by all those calls together. The names are internal \pytorch names for gradient computing functions. Measurements were taken on NVIDIA GeForce RTX 3090 24GB GPU.}
    \label{tab:node-timings}
\end{table}

\subsection{Memory}

Another advantage of the REINFORCE estimator is that it utilizes less memory. This is again due to the workings of the \texttt{torch.autograd}, the module responsible for automatic differentiation.  As mentioned before, the computational graph created during the forward pass stores in each node partial results required for gradient computation on the backward pass. Using the hook mechanism of \pytorch we have measured the number of different tensors stored in the graph and their total size. The results are presented in Table~\ref{tab:memory},  $M_1$ is the total amount of memory taken by tensors stored in the computational graph, $M_2$ is the memory usage as reported by the {\tt torch.cuda.max\_memory\_allocated} function and n.t. is the number of different tensors stored in the graph. We can see that in the case of the REINFORCE algorithm, the number of tensors does not grow with the lattice size, contrary to the reparameterization trick. This is consistent with what we have said already about the DAG in the previous section. The difference in memory usage is of the order of $20\%$ for $L=16$ and over $50\%$ for $L=24$. The numbers reported in table~\ref{tab:memory} are only a lower bound on the memory usage, in practice total allocated memory as reported by \texttt{nvidia-smi} utility can be substantially higher. This reduction in the allocated memory translates directly into the gains in speed. This can be seen in table~\ref{tab:timings}, lower memory usage resulted in larger batches fitting on the GPU which in turn resulted in faster gradient evaluation. 

\begin{table}
    \centering
    \begin{tabular}{|rr|rrr| rrr|}
    \hline\hline
    L &batch &\multicolumn{3}{c|}{r.t.} &  \multicolumn{3}{c|}{REINF.}\\\hline\hline
    & &$M_1$& $M_2$ & n.t.  & $M_1$ & $M_2$ & n.t. \\  \hline
     16 & 128   & 3.09 & 3.48&  5296 & 2.6 & 2.76 &  3401 \\
     16 & 256   & 6.16 & 6.87&  5292 &5.19  & 5.45&  3401 \\ 
     24 & 128   & 7.88& 9.41 &  7847  & 5.37 & 5.65  &3401\\
     24 & 256   & 15.72 & 18.72  & 7827 & 10.74  & 11.25  & 3401 \\
         \hline\hline
    \end{tabular}
    \caption{GPU memory  (in GB)  allocated during a single call to loss function, $M_1$ is the total amount of memory taken by tensors stored in computational graph, $M_2$ is the memory usage as reported by the {\tt torch.cuda.max\_memory\_allocated} function and n.t. is the number of different tensors stored in the graph.}
    \label{tab:memory}
\end{table}

\section{Summary}

In this paper, we have advocated the use of the REINFORCE type gradient estimator while training the normalizing flows. The advantage of this estimator is that it avoids calculating gradients of the target distribution which may be beneficial if this distribution is computationally intensive. We have applied this estimator to the 2D Schwinger model with two flavours of Wilson fermions which has an action in the form of the determinant of the Dirac operator. We have found that this estimator has better convergence properties, at least for the parameters used in this study, and is more numerically stable. We demonstrated that it is much faster and takes less memory.

The  2D Schwinger model investigated here can be considered as a "toy model" of lattice QCD, sharing with it the fermionic action containing the determinant of the Dirac operator. However the direct calculation of this determinant cannot be scaled up to any useful lattice sizes. In such cases, the determinant is estimated by introducing additional so-called pseudo-fermions fields. Such an approach was also implemented using normalizing flows in Ref.~\cite{AbbotPseudoFermions} where the Schwinger model was also considered (see also \cite{AlbergoFermionicLatticeTheories}).  It is hard to predict the relative performance of the \reinf{} and 
 \rt{} estimators in that case. We have found that the slowing down of the \rt{} gradient estimator comes mainly from the assembly of the Dirac operator, rather than from the calculation of the determinant itself. That suggests that it can be highly implementation dependent. Checking if the \reinf{} estimator retains its advantage over the \rt{} with pseudo-fermions is a subject of ongoing work.

We would also like  to check if those benefits can be obtained for other models, especially those outside of the lattice field theory domain. A comparison with other than \rt{} estimators, notably path gradients \cite{pathGradient}, is also in order. This would be the subject of further work.

\section*{Acknowledgment}
Computer time allocation grant \texttt{plnglft} on the Ares and Athena supercomputers hosted by AGH Cyfronet in Kraków, Poland were used through the Polish PLGRID consortium. T.S. kindly acknowledges the support of the Polish National Science Center (NCN) Grants No.\,2019/32/C/ST2/00202 and 2021/43/D/ST2/03375 and support of the Faculty of Physics, Astronomy and Applied Computer Science, Jagiellonian University Grant No.~LM/23/ST. This research was partially funded by the Priority Research Area Digiworld under the program Excellence Initiative – Research University at the Jagiellonian University in Kraków.

\appendix

\section{Implementation}
\label{app:implementation}

As mentioned before we have implemented the same model architecture as in the Ref.~\cite{PhialaSchwinger}. The source code for our implementation can be found in \cite{code}. We have used the gauge-equivariant coupling layers with each layer updating a subset of links given by 
\begin{equation}
    M^k_{\mu\nu} = 
    \{
    U_\mu((4n+k)\hat\mu + 2m\hat\nu)|\ \forall n,m\in \mathbb{Z}
    \}
    \cup
     \{
    U_\mu((4n+2+2k)\hat\mu + (2m+1)\hat\nu)|\ \forall n,m\in \mathbb{Z}
    \}
\end{equation}
All links were updated in eight layers by first iterating over $k$ with $\mu=0,\nu=1$ and then again with with $\mu=1,\nu=0$ We used 48 layers so each link was updated six times. 

Plaquette couplings were given by the eight knots {\em circular splines} \cite{NeuralSplines,ToriSphere}.
The parameters of the splines were set by a neural network in each coupling layer. The network took as input not only the inactive plaquettes but also $2\times 1$ and $1\times 2$ Wilson loops. For $U(1)$ theory, each plaquette/loop was given by a single angle $\theta$ and we used the $(\cos\theta, \sin\theta)$ pairs as the input to the network.
The neural network was built from three convolutional layers with kernel size three and dilation factors $1,2,3$. We used 64 channels between convolutions and $\texttt{LeakyReLU}$ activation after each but the last layer. For more details please consult the Ref.~\cite{PhialaSchwinger} and/or the source code \cite{code}. 

The gradient estimators described in section \ref{sec:grad-estimators}  were implemented as {\em loss functions}. Each loss function took as the input a batch of generated prior configurations, model and the action and returned the overall loss on this batch, as well as the logarithms of the probabilities $\log q$ and $\log P$.  The returned loss could be used for automatic gradient calculations by calling \lstinline!.backward()! on it. Striped-down versions of the loss function for both estimators are presented in listings~\ref{lst:grt} and ~\ref{lst:gre}. Those loss functions could be subsequently used in a generic training step presented in listing~\ref{lst:train-step}, so the whole difference between the \rt{} and \reinf{} implementation was confined to those two functions.

\begin{lstlisting}[language=Python, float, caption={Loss function for  reparameterization based $\grt$ estimator.}, captionpos=b, numbers=none, numberstyle=\tiny, label={lst:grt}]
def grt_loss(z, log_prob_z, *, model, action, use_amp):
    layers = model['layers']

    with autocast(enabled=use_amp):
            x, logq = nf.apply_flow(layers, z, log_prob_z)
        
        logp = -action(x)
        loss = nf.calc_dkl(logp, logq)
        
        return loss, logq.detach(), logp.detach()
\end{lstlisting}

\begin{lstlisting}[language=Python, float, caption={Loss function for the REINFORCE based $\gri$ estimator.}, captionpos=b, numbers=none, numberstyle=\tiny, label={lst:gre}]
def gre_loss(sub_mean, z_a, log_prob_z_a, *,
            model, action, use_amp):
    layers, prior = model['layers'], model['prior']
    with torch.no_grad():
        with autocast(enabled=use_amp):
            phi, logq = nf.apply_flow(layers, z_a, log_prob_z_a)
            logp = -action(phi)
            signal = logq - logp

    with autocast(enabled=use_amp):
        z, log_q_phi = nf.reverse_apply_flow(layers, phi, 
            torch.zeros_like(log_prob_z_a, device=phi.device))
        prob_z = prior.log_prob(z)
        log_q_phi = prob_z - log_q_phi
        loss = torch.mean(log_q_phi * (signal - signal.mean()))
      

    return loss, logq, logp
\end{lstlisting}

\begin{lstlisting}[float, caption={Single training step. It accumulates gradient over \texttt{n\_batches} of size \texttt{batch\_size}. The difference between different gradient estimators is encapsulated in the loss function \texttt{loss\_fn}. }
, captionpos=b, numbers=none, numberstyle=\tiny, label={lst:train-step}, language=Python]

def train_step(*, model, action, loss_fn, batch_size, 
             optimizer, scheduler=None, n_batches=1, use_amp):
    optimizer.zero_grad(set_to_none=True)
    prior = model['prior']
    
    for i in range(n_batches):
        with autocast(enabled=use_amp):
            z = prior.sample_n(batch_size=batch_size)
            log_prob_z = prior.log_prob(z)
       
        l, logq, logp = loss_fn(z, log_prob_z, 
            model=model, action=action, use_amp=use_amp)
    
        l.backward()
    
    # Gradient clipping here  

    optimizer.step()
    if scheduler is not None:
        scheduler.step()
\end{lstlisting}

\bibliography{nmcmc}

\begin{thebibliography}{29}
\expandafter\ifx\csname natexlab\endcsname\relax\def\natexlab#1{#1}\fi
\providecommand{\url}[1]{\texttt{#1}}
\providecommand{\href}[2]{#2}
\providecommand{\path}[1]{#1}
\providecommand{\DOIprefix}{doi:}
\providecommand{\ArXivprefix}{arXiv:}
\providecommand{\URLprefix}{URL: }
\providecommand{\Pubmedprefix}{pmid:}
\providecommand{\doi}[1]{\href{http://dx.doi.org/#1}{\path{#1}}}
\providecommand{\Pubmed}[1]{\href{pmid:#1}{\path{#1}}}
\providecommand{\bibinfo}[2]{#2}
\ifx\xfnm\relax \def\xfnm[#1]{\unskip,\space#1}\fi
\bibitem[{Metropolis et~al.(1953)Metropolis, Rosenbluth, Rosenbluth, Teller,
  and Teller}]{metropolis}
\bibinfo{author}{N.~Metropolis}, \bibinfo{author}{A.~W. Rosenbluth},
  \bibinfo{author}{M.~N. Rosenbluth}, \bibinfo{author}{A.~H. Teller},
  \bibinfo{author}{E.~Teller},
\newblock \bibinfo{title}{Equation of state calculations by fast computing
  machines},
\newblock \bibinfo{journal}{The Journal of Chemical Physics}
  \bibinfo{volume}{21} (\bibinfo{year}{1953}) \bibinfo{pages}{1087--1092}.
  \DOIprefix\doi{10.1063/1.1699114}.
\bibitem[{Hastings(1970)}]{Hastings}
\bibinfo{author}{W.~K. Hastings},
\newblock \bibinfo{title}{{Monte Carlo sampling methods using Markov chains and
  their applications}},
\newblock \bibinfo{journal}{Biometrika} \bibinfo{volume}{57}
  (\bibinfo{year}{1970}) \bibinfo{pages}{97--109}.
  \DOIprefix\doi{10.1093/biomet/57.1.97}.
\bibitem[{Albergo et~al.(2019)Albergo, Kanwar, and
  Shanahan}]{PhysRevD.100.034515}
\bibinfo{author}{M.~S. Albergo}, \bibinfo{author}{G.~Kanwar},
  \bibinfo{author}{P.~E. Shanahan},
\newblock \bibinfo{title}{Flow-based generative models for markov chain monte
  carlo in lattice field theory},
\newblock \bibinfo{journal}{Phys. Rev. D} \bibinfo{volume}{100}
  (\bibinfo{year}{2019}) \bibinfo{pages}{034515}.
\bibitem[{Wu et~al.(2019)Wu, Wang, and Zhang}]{VANPRL}
\bibinfo{author}{D.~Wu}, \bibinfo{author}{L.~Wang}, \bibinfo{author}{P.~Zhang},
\newblock \bibinfo{title}{Solving statistical mechanics using variational
  autoregressive networks},
\newblock \bibinfo{journal}{Phys. Rev. Lett.} \bibinfo{volume}{122}
  (\bibinfo{year}{2019}) \bibinfo{pages}{080602}.
\bibitem[{Nicoli et~al.(2020)Nicoli, Nakajima, Strodthoff, Samek, M\"uller, and
  Kessel}]{PhysRevE.101.023304}
\bibinfo{author}{K.~A. Nicoli}, \bibinfo{author}{S.~Nakajima},
  \bibinfo{author}{N.~Strodthoff}, \bibinfo{author}{W.~Samek},
  \bibinfo{author}{K.-R. M\"uller}, \bibinfo{author}{P.~Kessel},
\newblock \bibinfo{title}{Asymptotically unbiased estimation of physical
  observables with neural samplers},
\newblock \bibinfo{journal}{Phys. Rev. E} \bibinfo{volume}{101}
  (\bibinfo{year}{2020}) \bibinfo{pages}{023304}.
\bibitem[{Bia\l{}as et~al.(2023)Bia\l{}as, Korcyl, and Stebel}]{Bialas:2021bei}
\bibinfo{author}{P.~Bia\l{}as}, \bibinfo{author}{P.~Korcyl},
  \bibinfo{author}{T.~Stebel},
\newblock \bibinfo{title}{{Analysis of autocorrelation times in neural Markov
  chain Monte Carlo simulations}},
\newblock \bibinfo{journal}{Phys. Rev. E} \bibinfo{volume}{107}
  (\bibinfo{year}{2023}) \bibinfo{pages}{015303}.
  \DOIprefix\doi{10.1103/PhysRevE.107.015303}.
  \href{http://arxiv.org/abs/2111.10189}{{\tt arXiv:2111.10189}}.
\bibitem[{Bia\l{}as et~al.(2022)Bia\l{}as, Korcyl, and Stebel}]{Bialas:2022qbs}
\bibinfo{author}{P.~Bia\l{}as}, \bibinfo{author}{P.~Korcyl},
  \bibinfo{author}{T.~Stebel},
\newblock \bibinfo{title}{{Hierarchical autoregressive neural networks for
  statistical systems}},
\newblock \bibinfo{journal}{Comput. Phys. Commun.} \bibinfo{volume}{281}
  (\bibinfo{year}{2022}) \bibinfo{pages}{108502}.
  \DOIprefix\doi{10.1016/j.cpc.2022.108502}.
  \href{http://arxiv.org/abs/2203.10989}{{\tt arXiv:2203.10989}}.
\bibitem[{Bia\l{}as et~al.(2023)Bia\l{}as, Czarnota, Korcyl, and
  Stebel}]{Bialas:2022bdl}
\bibinfo{author}{P.~Bia\l{}as}, \bibinfo{author}{P.~Czarnota},
  \bibinfo{author}{P.~Korcyl}, \bibinfo{author}{T.~Stebel},
\newblock \bibinfo{title}{{Simulating first-order phase transition with
  hierarchical autoregressive networks}},
\newblock \bibinfo{journal}{Phys. Rev. E} \bibinfo{volume}{107}
  (\bibinfo{year}{2023}) \bibinfo{pages}{054127}.
  \DOIprefix\doi{10.1103/PhysRevE.107.054127}.
  \href{http://arxiv.org/abs/2212.04955}{{\tt arXiv:2212.04955}}.
\bibitem[{Białas et~al.(2023)Białas, Korcyl, and Stebel}]{Bialas:2023fjz}
\bibinfo{author}{P.~Białas}, \bibinfo{author}{P.~Korcyl},
  \bibinfo{author}{T.~Stebel}, \bibinfo{title}{Mutual information of spin
  systems from autoregressive neural networks}, \bibinfo{year}{2023}.
  \href{http://arxiv.org/abs/2304.13412}{{\tt arXiv:2304.13412}}.
\bibitem[{Kingma and Welling(2013)}]{reparameterization}
\bibinfo{author}{D.~P. Kingma}, \bibinfo{author}{M.~Welling},
\newblock \bibinfo{title}{Auto-encoding variational bayes},
\newblock \bibinfo{journal}{2nd International Conference on Learning
  Representations, ICLR 2014 - Conference Track Proceedings}
  (\bibinfo{year}{2013}). \URLprefix \url{https://arxiv.org/abs/1312.6114v11}.
\bibitem[{Bialas et~al.(2022)Bialas, Korcyl, and Stebel}]{bialas2022gradient}
\bibinfo{author}{P.~Bialas}, \bibinfo{author}{P.~Korcyl},
  \bibinfo{author}{T.~Stebel}, \bibinfo{title}{Gradient estimators for
  normalising flows}, \bibinfo{year}{2022}.
  \href{http://arxiv.org/abs/2202.01314}{{\tt arXiv:2202.01314}}.
\bibitem[{Vaitl et~al.(2022)Vaitl, Nicoli, Nakajima, and Kessel}]{pathGradient}
\bibinfo{author}{L.~Vaitl}, \bibinfo{author}{K.~A. Nicoli},
  \bibinfo{author}{S.~Nakajima}, \bibinfo{author}{P.~Kessel},
\newblock \bibinfo{title}{Gradients should stay on path: better estimators of
  the reverse- and forward kl divergence for normalizing flows},
\newblock \bibinfo{journal}{Machine Learning: Science and Technology}
  \bibinfo{volume}{3} (\bibinfo{year}{2022}) \bibinfo{pages}{045006}.
  \DOIprefix\doi{10.1088/2632-2153/ac9455}.
\bibitem[{Albergo et~al.(2022)Albergo, Boyda, Cranmer, Hackett, Kanwar,
  Racani\`ere, Rezende, Romero-L\'opez, Shanahan, and Urban}]{PhialaSchwinger}
\bibinfo{author}{M.~S. Albergo}, \bibinfo{author}{D.~Boyda},
  \bibinfo{author}{K.~Cranmer}, \bibinfo{author}{D.~C. Hackett},
  \bibinfo{author}{G.~Kanwar}, \bibinfo{author}{S.~Racani\`ere},
  \bibinfo{author}{D.~J. Rezende}, \bibinfo{author}{F.~Romero-L\'opez},
  \bibinfo{author}{P.~E. Shanahan}, \bibinfo{author}{J.~M. Urban},
\newblock \bibinfo{title}{Flow-based sampling in the lattice schwinger model at
  criticality},
\newblock \bibinfo{journal}{Phys. Rev. D} \bibinfo{volume}{106}
  (\bibinfo{year}{2022}) \bibinfo{pages}{014514}.
  \DOIprefix\doi{10.1103/PhysRevD.106.014514}.
\bibitem[{Białas et~al.(2023)Białas, Korcyl, and Stebel}]{code}
\bibinfo{author}{P.~Białas}, \bibinfo{author}{P.~Korcyl},
  \bibinfo{author}{T.~Stebel}, \bibinfo{year}{2023}. \URLprefix
  \url{https://github.com/nmcmc/nmcmc-code.git}.
\bibitem[{Liu(1996)}]{Liu}
\bibinfo{author}{J.~S. Liu},
\newblock \bibinfo{title}{Metropolized independent sampling with comparisons to
  rejection sampling and importance sampling},
\newblock \bibinfo{journal}{Statistics and Computing} \bibinfo{volume}{6}
  (\bibinfo{year}{1996}) \bibinfo{pages}{113--119}.
\bibitem[{Kullback and Leibler(1951)}]{KL}
\bibinfo{author}{S.~Kullback}, \bibinfo{author}{R.~A. Leibler},
\newblock \bibinfo{title}{{On Information and Sufficiency}},
\newblock \bibinfo{journal}{The Annals of Mathematical Statistics}
  \bibinfo{volume}{22} (\bibinfo{year}{1951}) \bibinfo{pages}{79--86}.
  \DOIprefix\doi{10.1214/aoms/1177729694}.
\bibitem[{Nicoli et~al.(2021)Nicoli, Anders, Funcke, Hartung, Jansen, Kessel,
  Nakajima, and Stornati}]{PhysRevLett.126.032001}
\bibinfo{author}{K.~A. Nicoli}, \bibinfo{author}{C.~J. Anders},
  \bibinfo{author}{L.~Funcke}, \bibinfo{author}{T.~Hartung},
  \bibinfo{author}{K.~Jansen}, \bibinfo{author}{P.~Kessel},
  \bibinfo{author}{S.~Nakajima}, \bibinfo{author}{P.~Stornati},
\newblock \bibinfo{title}{Estimation of thermodynamic observables in lattice
  field theories with deep generative models},
\newblock \bibinfo{journal}{Phys. Rev. Lett.} \bibinfo{volume}{126}
  (\bibinfo{year}{2021}) \bibinfo{pages}{032001}.
  \DOIprefix\doi{10.1103/PhysRevLett.126.032001}.
\bibitem[{Dinh et~al.(2017)Dinh, Sohl-Dickstein, and Bengio}]{dinh2017density}
\bibinfo{author}{L.~Dinh}, \bibinfo{author}{J.~Sohl-Dickstein},
  \bibinfo{author}{S.~Bengio}, \bibinfo{title}{Density estimation using real
  nvp}, \bibinfo{year}{2017}. \href{http://arxiv.org/abs/1605.08803}{{\tt
  arXiv:1605.08803}}.
\bibitem[{Kobyzev et~al.(2020)Kobyzev, Prince, and Brubaker}]{9089305}
\bibinfo{author}{I.~Kobyzev}, \bibinfo{author}{S.~Prince},
  \bibinfo{author}{M.~Brubaker},
\newblock \bibinfo{title}{Normalizing flows: An introduction and review of
  current methods},
\newblock \bibinfo{journal}{IEEE Transactions on Pattern Analysis and Machine
  Intelligence}  (\bibinfo{year}{2020}) \bibinfo{pages}{1--1}.
\bibitem[{Paszke et~al.(2019)}]{PyTorch}
\bibinfo{author}{A.~Paszke}, et~al.,
\newblock \bibinfo{title}{Pytorch: An imperative style, high-performance deep
  learning library},
\newblock in: \bibinfo{booktitle}{Advances in Neural Information Processing
  Systems 32}, \bibinfo{publisher}{Curran Associates, Inc.},
  \bibinfo{year}{2019}, pp. \bibinfo{pages}{8024--8035}.
\bibitem[{Baydin et~al.(2018)Baydin, Pearlmutter, Radul, and Siskind}]{AD}
\bibinfo{author}{A.~G. Baydin}, \bibinfo{author}{B.~A. Pearlmutter},
  \bibinfo{author}{A.~A. Radul}, \bibinfo{author}{J.~M. Siskind},
\newblock \bibinfo{title}{Automatic differentiation in machine learning: a
  survey},
\newblock \bibinfo{journal}{Journal of Machine Learning Research}
  \bibinfo{volume}{18} (\bibinfo{year}{2018}) \bibinfo{pages}{1--43}.
  \URLprefix \url{http://jmlr.org/papers/v18/17-468.html}.
\bibitem[{Mnih and Rezende(2016)}]{Rezende2016}
\bibinfo{author}{A.~Mnih}, \bibinfo{author}{D.~J. Rezende},
\newblock \bibinfo{title}{Variational inference for monte carlo objectives},
\newblock in: \bibinfo{booktitle}{Proceedings of the 33rd International
  Conference on International Conference on Machine Learning - Volume 48},
  ICML'16, \bibinfo{publisher}{JMLR.org}, \bibinfo{year}{2016}, p.
  \bibinfo{pages}{2188–2196}.
\bibitem[{Kanwar et~al.(2020)Kanwar, Albergo, Boyda, Cranmer, Hackett,
  Racani\`ere, Rezende, and Shanahan}]{PhialaEquivariant}
\bibinfo{author}{G.~Kanwar}, \bibinfo{author}{M.~S. Albergo},
  \bibinfo{author}{D.~Boyda}, \bibinfo{author}{K.~Cranmer},
  \bibinfo{author}{D.~C. Hackett}, \bibinfo{author}{S.~Racani\`ere},
  \bibinfo{author}{D.~J. Rezende}, \bibinfo{author}{P.~E. Shanahan},
\newblock \bibinfo{title}{Equivariant flow-based sampling for lattice gauge
  theory},
\newblock \bibinfo{journal}{Phys. Rev. Lett.} \bibinfo{volume}{125}
  (\bibinfo{year}{2020}) \bibinfo{pages}{121601}.
  \DOIprefix\doi{10.1103/PhysRevLett.125.121601}.
\bibitem[{Albergo et~al.(2021)}]{albergo2021introduction}
\bibinfo{author}{M.~S. Albergo}, et~al., \bibinfo{title}{Introduction to
  normalizing flows for lattice field theory}, \bibinfo{year}{2021}.
  \href{http://arxiv.org/abs/2101.08176}{{\tt arXiv:2101.08176}}.
\bibitem[{Durkan et~al.(2019)Durkan, Bekasov, Murray, and
  Papamakarios}]{NeuralSplines}
\bibinfo{author}{C.~Durkan}, \bibinfo{author}{A.~Bekasov},
  \bibinfo{author}{I.~Murray}, \bibinfo{author}{G.~Papamakarios},
\newblock \bibinfo{title}{Neural spline flows},
\newblock \bibinfo{journal}{33rd Conference on Neural Information Processing
  Systems (NeurIPS 2019), Vancouver, Canada}  (\bibinfo{year}{2019}).
  \DOIprefix\doi{10.48550/arxiv.1906.04032}.
\bibitem[{Rezende et~al.(2020)Rezende, Papamakarios, Racanière, Albergo,
  Kanwar, Shanahan, and Cranmer}]{ToriSphere}
\bibinfo{author}{D.~J. Rezende}, \bibinfo{author}{G.~Papamakarios},
  \bibinfo{author}{S.~Racanière}, \bibinfo{author}{M.~S. Albergo},
  \bibinfo{author}{G.~Kanwar}, \bibinfo{author}{P.~E. Shanahan},
  \bibinfo{author}{K.~Cranmer},
\newblock \bibinfo{title}{Normalizing flows on tori and spheres},
\newblock \bibinfo{journal}{37th International Conference on Machine Learning,
  ICML 2020}  (\bibinfo{year}{2020}) \bibinfo{pages}{8039--8048}.
  \DOIprefix\doi{10.48550/arxiv.2002.02428}.
\bibitem[{PyTorch(2023)}]{autograd}
\bibinfo{author}{PyTorch}, \bibinfo{title}{A gentle introduction to
  \texttt{torch.autograd}}, \bibinfo{year}{2023}. \URLprefix
  \url{https://pytorch.org/tutorials/beginner/blitz/autograd\_tutorial.html}.
\bibitem[{Abbott et~al.(2022)Abbott, Albergo, Boyda, Cranmer, Hackett, Kanwar,
  Racani\`ere, Rezende, Romero-L\'opez, Shanahan, Tian, and
  Urban}]{AbbotPseudoFermions}
\bibinfo{author}{R.~Abbott}, \bibinfo{author}{M.~S. Albergo},
  \bibinfo{author}{D.~Boyda}, \bibinfo{author}{K.~Cranmer},
  \bibinfo{author}{D.~C. Hackett}, \bibinfo{author}{G.~Kanwar},
  \bibinfo{author}{S.~Racani\`ere}, \bibinfo{author}{D.~J. Rezende},
  \bibinfo{author}{F.~Romero-L\'opez}, \bibinfo{author}{P.~E. Shanahan},
  \bibinfo{author}{B.~Tian}, \bibinfo{author}{J.~M. Urban},
\newblock \bibinfo{title}{Gauge-equivariant flow models for sampling in lattice
  field theories with pseudofermions},
\newblock \bibinfo{journal}{Phys. Rev. D} \bibinfo{volume}{106}
  (\bibinfo{year}{2022}) \bibinfo{pages}{074506}. \URLprefix
  \url{https://link.aps.org/doi/10.1103/PhysRevD.106.074506}.
  \DOIprefix\doi{10.1103/PhysRevD.106.074506}.
\bibitem[{Albergo et~al.(2021)Albergo, Kanwar, Racani\`ere, Rezende, Urban,
  Boyda, Cranmer, Hackett, and Shanahan}]{AlbergoFermionicLatticeTheories}
\bibinfo{author}{M.~S. Albergo}, \bibinfo{author}{G.~Kanwar},
  \bibinfo{author}{S.~Racani\`ere}, \bibinfo{author}{D.~J. Rezende},
  \bibinfo{author}{J.~M. Urban}, \bibinfo{author}{D.~Boyda},
  \bibinfo{author}{K.~Cranmer}, \bibinfo{author}{D.~C. Hackett},
  \bibinfo{author}{P.~E. Shanahan},
\newblock \bibinfo{title}{Flow-based sampling for fermionic lattice field
  theories},
\newblock \bibinfo{journal}{Phys. Rev. D} \bibinfo{volume}{104}
  (\bibinfo{year}{2021}) \bibinfo{pages}{114507}.
  \DOIprefix\doi{10.1103/PhysRevD.104.114507}.

\end{thebibliography}
\end{document}